\documentclass[sigconf]{aamas} 

\usepackage{balance} 

\setcopyright{ifaamas}
\acmConference[AAMAS '24]{Proc.\@ of the 23rd International Conference
on Autonomous Agents and Multiagent Systems (AAMAS 2024)}{May 6 -- 10, 2024}
{Auckland, New Zealand}{N.~Alechina, V.~Dignum, M.~Dastani, J.S.~Sichman (eds.)}
\copyrightyear{2024}
\acmYear{2024}
\acmDOI{}
\acmPrice{}
\acmISBN{}

\acmSubmissionID{741}

\title[AAMAS-2024 Formatting Instructions]{BrainSLAM: SLAM on Neural Population Activity Data
}

\author{Kipp Freud}
\affiliation{
  \institution{University of Bristol}
  \city{Bristol}
  \country{United Kingdom}}
\email{km14740@bristol.ac.uk}

\author{Nathan Lepora}
\affiliation{
  \institution{University of Bristol}
  \city{Bristol}
  \country{United Kingdom}}
\email{n.lepora@bristol.ac.uk}

\author{Matt W. Jones}
\affiliation{
  \institution{University of Bristol}
  \city{Bristol}
  \country{United Kingdom}}
\email{Matt.Jones@bristol.ac.uk}

\author{Cian O'Donnell}
\affiliation{
  \institution{Ulster University}
  \city{Derry/Londonderry}
  \country{United Kingdom}}
\email{C.ODonnell2@Ulster.ac.uk}

\begin{abstract}
Simultaneous localisation and mapping (SLAM) algorithms are commonly used in robotic systems for learning maps of novel environments. Brains also appear to learn maps, but the mechanisms are not known and it is unclear how to infer these maps from neural activity data. We present BrainSLAM; a method for performing SLAM using only population activity (local field potential, LFP) data simultaneously recorded from three brain regions in rats: hippocampus, prefrontal cortex, and parietal cortex. This system uses a convolutional neural network (CNN) to decode velocity and familiarity information from wavelet scalograms of neural local field potential data recorded from rats as they navigate a 2D maze. The CNN's output drives a RatSLAM-inspired architecture, powering an attractor network which performs path integration plus a separate system which performs `loop closure' (detecting previously visited locations and correcting map aliasing errors). Together, these three components can construct faithful representations of the environment while simultaneously tracking the animal's location. This is the first demonstration of inference of a spatial map from brain recordings. Our findings expand SLAM to a new modality, enabling a new method of mapping environments and facilitating a better understanding of the role of cognitive maps in navigation and decision making.
\end{abstract}

\keywords{Neural Decoding; Mapping; SLAM; RatSLAM; Deep Learning; Biological Robots}

\newcommand{\BibTeX}{\rm B\kern-.05em{\sc i\kern-.025em b}\kern-.08em\TeX}

\begin{document}

\pagestyle{fancy}
\fancyhead{}

\maketitle 

\section{Introduction}

Cognitive maps can be conceived as spatial representations within the brain that are derived across any number of individual experiences. Such spatial representaitons would include information about paths not directly traveled but that can be inferred from the relationships among experienced routes \cite{kitchin2001cognitive}. Once a cognitive map is formed, it can be reactivated when an animal subsequently enters the same environment and updated using information from new experiences that occur in that environment. When familiar routes to a goal are blocked, use of the cognitive map enables navigation through alternative paths because information about novel (i.e. never before traveled) routes is included in a single representational structure of the environment. 

Evidence supporting the cognitive map hypothesis was presented in \cite{o1971hippocampus}. Here, researchers recorded rodent hippocampal activity as these animals moved freely in an environment. Analysis of the gathered data showed the presence of place cells; hippocampal cells with firing rates modulated by the location of the animal. The firing rates of each cell increased dramatically when the animal was moving through a specific region of an environment, known as the place field of that cell \cite{o1978hippocampus}. Further research has discovered many other types of spatially sensitive neuron which are believed to form the basis of the cognitive map \cite{grieves2017representation}.

Though the concept of an internal cognitive map has been explored in much detail \cite{schiller2015memory}, no algorithms have yet been presented which attempt to infer an explicit representation of the internal cognitive map from individual animals or people. The extraction of such representations would be valuable for two reasons. First, these representations would be useful for both neurobiology research and bio-inspired control. For example, we could examine to what extent the navigational decisions an animal makes as it moves through an environment can be predicted by its inferred cognitive map of that environment. This would help us understand if and how animals use cognitive maps for navigation, test how map use varies depending on context, and design control algorithms based on these observations. Secondly, methods which extracted cognitive maps could be used within brain-computer interfaces (BCIs) to provide an innovative approach for humans to map out their surroundings. By decoding cognitive maps from neural data, we could better understand how we interact with and perceive our spaces, offering applications in areas like virtual reality, augmented reality, and even physical navigation assistance for those with impairments.

Simultaneous localisation and mapping (SLAM) is the problem of using an agent to build a map of an unknown environment while at the same time maintaining a location estimate of that agent as it moves through that environment. There are many proposed solutions to the SLAM problem \cite{aulinas2008slam}, the majority of which can be summarized as methods for taking noisy data encoding velocity and location information and generating maps.

The goal of this project was to use local field potential (LFP) data recorded from the brains of rats as they navigated a 2D maze to perform SLAM. LFP is a measure of the spatially-weighted average of electrical potential in the extracellular space around populations of neurons and their synaptic connections \cite{einevoll2013local}. This data is gathered using intracranial tetrodes implanted in the hippocampus. Notably, LFP data taken from the hippocampus, prefrontal cortex, and parietal cortex, such as that used in the research presented here, has all been shown to encode velocity and location information to some degree \cite{o1998place, small2003posterior, colby1996spatial}; thus, one might expect that some SLAM algorithms may be extendable to this form of input. 

In this work, we propose a novel approach that facilitates the use of LFP data as an input to SLAM systems. Our proposed system uses a deep convolution neural network (CNN) to predict velocity and location information from LFP data \cite{frey2020interpreting}, and uses these predictions to power a RatSLAM architecture \cite{milford2004ratslam} which produces graphical representations of the environment being explored. Predictions from the CNN are used to transform activation within a 3-dimensional attractor network encoding position within an environment; path integration is performed on these estimated positions and mapped to a corresponding experience map representing our current knowledge of the environment being explored. A separate system uses location features extracted from the convolutional network to infer uniqueness of the current location and detect when the rodent has returned to a previously visited location; at which point, discrepancies between current estimated location and this previously visited location will be minimized in our experience map representation, preventing aliasing in the system. 

We provide visualisations of the inferred cognitive maps, showing that our system is capable of outputting high fidelity graphical maps given only LFP data. Such visualisations have never been produced before, and represent a new path to researching how cognitive maps are formed in the brain, which brain regions are necessary, and how these maps reflect experience and affect consequent behaviour.

\section{Methods}

\begin{figure*}[t]
  \centering
  \includegraphics[width=1\linewidth]{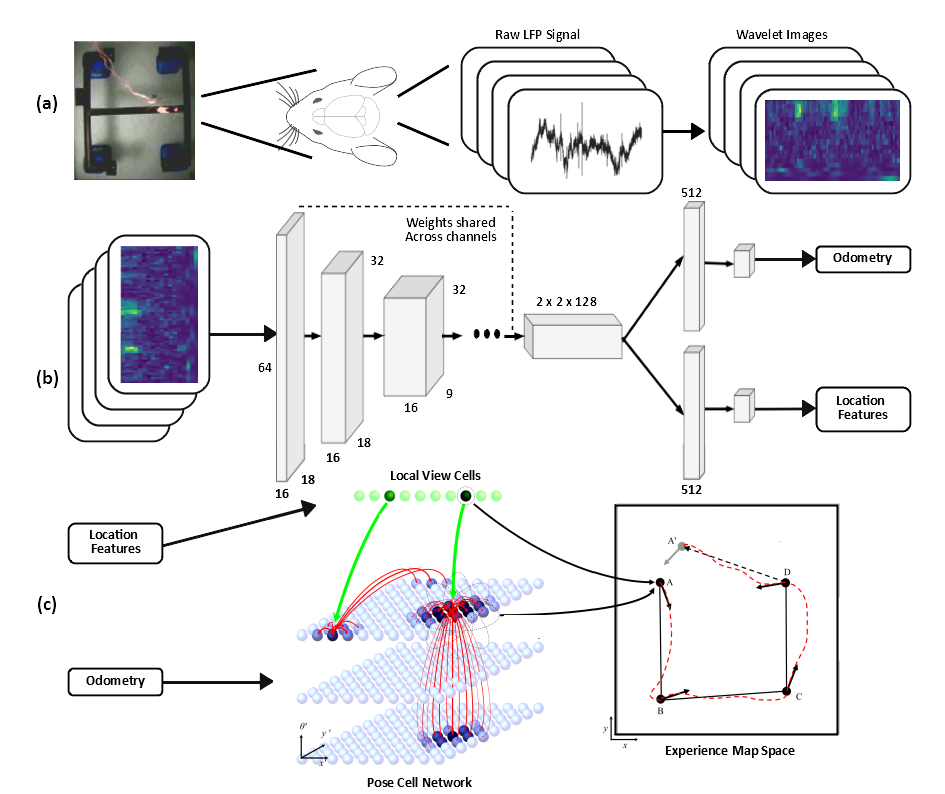}
  \caption{In the research presented here, local field potential data collected from the dorsal CA1 of hippocampus (HPC), prefrontal cortex (PFC), and the parietal cortex (PC) of rodents as they completed a simple behavioural task within a maze is transformed into wavelet decomposition images (a). This data is used to train a deep neural decoding system with 13 downsampling layers and separate fully connected layers for each output; this network predicts both odometry values (speed and direction) and location (b). These predictions are used to shift activation within a competitive 3D attractor network, powering a RatSLAM architecture which produces graphical representations of the environment being explored (c). Panel c adapted with permission from \citet{milford2010persistent}}
  \label{fig:bigfig}
  \vspace{-1em}
\end{figure*}

\subsection{Dataset \& Data Preprocessing}

The data for this project was recorded from the brains of 3 adult, male Long-Evans rats navigating a memory and decision-making task to find sucrose rewards on a maze \cite{jones2005theta}. We note that all procedures were performed from 2009 - 2010, and in accordance with the UK Animals Scientific Procedures Act (1986) and approved by the University of Bristol Animal Welfare and Ethical Review Board. This data has been utilized as part of numerous other studies \cite{freud2023deep, lloyd2012learning, jones2005theta, russo2023integration}.

The rats had to choose between left and right maze arms based on the direction of an initial guided turn (see Figure \ref{fig:bigfig}a). Animals were trained initially under a ``match turn'' rule (i.e. if initially forced to turn right, turn right again at the choice point, and vice versa). The maze was of dimension 1.3m $\times$ 1.7m, and this task was designed to incorporate both visual discrimination and a memory-guided responses. Rats ran 15–25 trials per 20- to 30-min recording session, and each trial of the task consisted of a `forced-turn' and `choice' epoch. 

While performing this task, data was collected from 16 chronically-implanted adjustable tetrodes placed in three distinct brain regions of each rat; dorsal CA1 of hippocampus (HPC), prefrontal cortex (PFC), and the parietal cortex (PC). LFP from each tetrode was sampled at 2kHz, bandpass filtered at 0.5-475Hz and reflects the aggregate activity of  populations of neurons and synapses near the tetrode tip \cite{maling2016local}. The animals' location, speed, and direction data was also calculated by tracking head-mounted LEDs at 25Hz. 

Head locations, as measured by LEDs, were first projected onto a skeleton representation of the maze to reduce noise caused by head movements that do not represent true movements through the maze -- for example, a rat moving their head from left to right as they travel down the central arm of the maze. This preprocessing step also simplified the direction estimation problem, as it reduced the search space of possible directions to only those which can be travelled on the skeleton representation (i.e. four directions, each $\pm$ 10 degrees of $0^{\circ}$, $90^{\circ}$, $180^{\circ}$, or $270^{\circ}$). Speed and direction values were calculated by examining position change across timesteps. Location, direction, and speed values are all used as outputs to train the deep neural decoding system described in Section \ref{ssec:di}. 

Following the approach described in \cite{frey2020interpreting}, raw LFP data was transformed using a wavelet decomposition to generate a three-dimensional representation depicting time, channels, and frequency bands \cite{torrence1998practical}. For the wavelet transformation, we used the Morlet wavelet:
\begin{equation}
\psi(t) = \pi^{-\frac{1}{4}} \exp(i\omega_0t ) \exp(-t^2/2),
\end{equation}
where $t$ is time and $\omega_0$ is a non-dimensional frequency constant (in our case $\omega_0 = 6$). Additionally, channel and frequency-wise normalization was applied using a median absolute deviation approach. Median and corresponding median absolute deviation values were calculated for each frequency and channel in our dataset, and inputs were then normalized using
\begin{align}X_i \leftarrow \frac{ X_i - \hat{X}_i }{\rm{median}(| X_i - \hat{X}_i |)},
\end{align}
where $\hat{X}_i$ denotes the mean activity for channel $i$.
Wavelet images were generated using 64 subsequent LFP values, meaning each image represented 32ms of recorded data. Thus for each 32ms window, a set of 16 wavelet images were generated: one for each tetrode present in the rodent. 

For each set of wavelet images generated, there exists in the dataset a corresponding ground truth location, speed, and direction value. A single maze session consisting of 20-30 minutes of activity was used for each rodent, with the first 80\% of the data collected for each rat used to train the deep neural decoding system described in Section \ref{ssec:di}, and the final 20\% used both as testing data and as input to our extended RatSLAM system to generate the maps shown in Section \ref{ssec:slamres}.

\subsection{Deep neural decoding for odometry and loop closure modules}
\label{ssec:di}

In \cite{frey2020interpreting}, a generalizable deep learning framework is presented for decoding sensory and behavioural variables directly from wide-band neural data. The approach requires little user input and generalizes across stimuli, behaviours, brain regions, and recording techniques. This framework has been shown to achieve state of the art results on neural localization tasks using Morlet wavelet decompositions of LFP recordings taken from the CA1 pyramidal cell layer in the hippocampus of freely foraging rodents, and is also capable of accurately decoding head direction and speed.  We note that this approach to decoding has been shown to be versatile to multiple forms of input including calcium imaging data \cite{pnevmatikakis2019analysis}, thus it is natural to assume that the method presented in this paper would also be versatile to these other modalities.   

A modified version of the architecture originally presented in \cite{frey2020interpreting} is used here. Our deep convolutional network uses only 13 downsampling time-distributed (i.e. with shared weights across channels) layers followed by two fully connected layers for each output. A kernel size of 3 was used throughout the model, and the number of filters was kept constant at 64 for the first 10 layers while sharing weights across the channel dimension, then doubling the number of filters for the following 3 layers. For downsampling the input, we used a stride of 2 intermixed between the time and frequency dimension. We used 2D-convolutions which share weights across the channel dimension for the first 10 convolutions and across the time dimension for the last 3 convolutional layers. Sharing weights across channels prevents overfitting to channel-specific features and thus improves generalisation by making sure there exists a global representation of important features e.g. of spikes or other prominent oscillations in the local field potential. The full architecture of the network used is presented in the supplemental material. 

Notably, as this system takes minimally-processed input, it is able to perform sensory decoding without spike sorting (a computationally-intensive process for detecting action potentials and assigning them to specific neurons \cite{lewicki1998review}).  Necessarily, spike sorting discards information in frequency bands outside of the spike range which potentially introduces biases implicit in the algorithm. Also, as the system is versatile to different forms of input, it is natural to think that the SLAM system presented here may also be extendable to non-LFP forms of neural input, for example two-photon calcium imaging data \cite{stosiek2003vivo}. 

Separate networks were trained for each rat using the training portion of the data generated from a single maze task session. Networks were trained to decode location, speed, and direction for 1000 epochs with 250 training steps per epoch. A batch size of 8 was used, as was an AMSGrad optimizer \cite{reddi2019convergence} with a learning rate of $7e^{-4}$. The decoding accuracy of the finished networks for each considered rat are shown and discussed in Section \ref{ssec:decode-results}.

Trained networks for each rat are used for two functions in our extended RatSLAM system: speed- and direction-decoding functionality is used as an odometry module in the system (discussed in Section \ref{sssec::vc}) and location decoding functionality is used as part of a loop closure module in the system to produce a similarity metric when comparing two units of input (discussed in Section \ref{sssec::Odom}).

\subsection{RatSLAM}

RatSLAM \citep{milford2004ratslam} is an approach to the SLAM problem inspired by computational models of the hippocampus of rodents; namely, the concept of place fields \citep{o1996geometric} and their modulation by the rodent's movement, visual stimuli or other sensory information. Though RatSLAM has only been shown to function using camera input, in principle it could be extended to accept any form of input that fulfils two conditions: first, it must be possible to infer velocity from input (or by comparing subsequent inputs); second, it must be possible to examine pairs of inputs and produce a familiarity score, giving a measure of confidence that the two input signals arose while the agent was in the same location. Here we present a modified RatSLAM system that achieves good results using only local field potential data. 

The RatSLAM system represents an agents pose by the activity in a competitive 3-dimensional attractor network called the pose cell network (Figure 1c). Path integration is performed by injecting activity into the pose cell network, shifting the current activity bump. New input data is compared with data associated with previously visited locations; if similarity is high, activation is injected into the particular pose cell associated with this previously visited location.

\subsubsection{Pose Cells}

The pose cell module is responsible for maintaining an estimate of the agent's position within the environment; it consists of a competitive 3-dimensional attractor network, with each node representing a point on a discretised version of the environment being explored, with the dimensions of the attractor network representing the $x$, $y$, and $ \theta $ coordinates. 

Each node in this network has strong excitatory connections to nearby nodes, and weaker inhibitory connections to more distal nodes. Global inhibition and normalisation also occurs at every time step; encapsulated by the following activation update rules, called in sequence at every time step:
\begin{flalign}
P_{ijk}^\prime  &\leftarrow  P_{ijk} + \sum_{a=0}^{N_x} \sum_{b=0}^{N_y} \sum_{c=0}^{N_\theta} {\epsilon}^{\rm{exc}}_{(a-i)(b-j)(c-k)} P_{abc}, 
\\
P_{ijk}^\prime  &\leftarrow P_{ijk} + \sum_{a=0}^{N_x} \sum_{b=0}^{N_y} \sum_{c=0}^{N_\theta} {\epsilon}^{\rm{inh}}_{(a-i)(b-j)(c-k)} P_{abc}, 
\\
P_{ijk} &\leftarrow \max(0, P_{ijk} - \psi), 
\\
P_{ijk}^\prime &\leftarrow \frac{P_{ijk}}{ \sum_{x=0}^{N_x} \sum_{y=0}^{N_y} \sum_{z=0}^{N_\theta} P_{xyz} }, 
\end{flalign}
where excitatory weights ${\epsilon}^{\rm{exc}}$ are generated using a three-dimensional discrete Gaussian distribution with variance 1, inhibitory weights ${\epsilon}^{\rm{inh}}$ are generated using a three dimensional discrete Gaussian distribution with variance 2, and $\psi$ is a global inhibition constant, in our case set to $2e^{-5}$. We also note that the $\theta$ dimension is wrapped, with nodes near the top of the $\theta$ dimension having connections to those at the bottom. 

Together these properties force the network's activity to converge to a single 3D bump. If activity is injected into the network far from the existing activity bump, a new bump will be created which competes with the original. If enough activity is injected into this new bump, it can become dominant and the old bump will disappear. 

We note here that activation is not only modulated by internal dynamics, but can also be injected into the network via excitatory connections from local view cells (discussed in Section \ref{sssec::vc}), and activation can be moved within the network via input from the odometry module (discussed in Section \ref{sssec::Odom}).

Location estimates are found by calculating the center of activation within this pose cell network: this is the mean of activation locations in the network, weighted by the activation at each location.

\subsubsection{View Cells}
\label{sssec::vc}

In the original RatSLAM system, local view cells each represent a distinct scene observed by the agent as it traverses the environment being explored. When new image data is processed, it is first examined by the view cell module to see if the data is similar enough to previously observed data. If similarity is above a certain threshold, the view cell corresponding to the most similar previously seen data will be activated, injecting activity into the pose cell network at the pose cell associated with the activated local view cell. If, on the other hand, all existing view cells correspond to data which is deemed dissimilar to the data being examined, a new view cell is added and is assigned excitatory connections to the most currently-activated pose cell. The excitatory links between view cells and pose cells represent the estimated position and direction of the agent when the corresponding data was observed.

In the system presented here, local view cells do not represent distinct visual scenes, but instead represent distinct periods of neural activity that are representative of particular locations within the maze. For example, while in a standard RatSLAM implementation a local view cell may be activated while the agent is observing a distinctive red door, in our system this cell may activate when the agent observes high frequency LFP oscillations coming from a certain tetrode.

Necessarily, we must provide a method to evaluate similarity between pieces of input data. In standard RatSLAM systems, this often makes use of an appearance-based view recognition system \citep{milford2004simultaneous}. In our system, however, we use the Euclidean distance between position estimates as predicted by the deep neural decoding system described in Section \ref{ssec:di} as a measure of difference; pieces of neural data which lead to similar decoded locations are likely to have been generated when the rat was in similar locations. If a new piece of data has a decoded positions within some distance (here 20 pixels; about 58mm) of a previously observed piece of data, the local view cell corresponding to the previously-observed piece of data will be activated. This process is called loop closure, and is instrumental for preventing error accumulation in the generated maps. 

When a local view cell is activated, activation is injected into the pose cell network before the internal dynamics of the network are applied at the location of the pose cell corresponding to the activated local view cell.

\subsubsection{Odometry}
\label{sssec::Odom}

The odometry module is used to examine new input and infer velocity. After velocity is estimated, activation is shifted in the pose cell network accordingly; i.e if a clockwise rotation of $60^\circ$ is predicted and the pose cell network's $\theta$ dimension is of size $360$, then all activation will be shifted 60 cells upwards, with any activation hitting the top of this dimension wrapping back to the bottom. Activation is shifted via stimulation from the odometry module at each time step after the internal dynamics of the attractor networks have been applied.

In a standard RatSLAM implementation, this is implemented by using a simple visual odometry system to compare two subsequent visual images captured by the agent using a visual odometry system \citep{nister2004visual}, yielding absolute speed and relative rotation between images. 

In the system presented here, however, absolute speed and rotation values are predicted by the deep neural decoding system described in Section \ref{ssec:di}. Relative rotation is found by calculating the difference between absolute rotation and current rotation estimate, as represented by the centre of activation of the pose cell network. Speed and inferred relative rotation are then used to shift activation in the pose cell network identically to a standard RatSLAM system \citep{milford2004ratslam}.

\subsubsection{Experience Map}

Activity in the pose cells and local view cells drives the creation of experiences. Experiences are represented by nodes in $(x,y,\theta)$ space connected by links representing transitions between experiences \citep{milford2007learning}. These nodes form the experience map; a graphical representation of the environment being explored. The first experience learned is initialised with an arbitrary (0,0,0) position within the experience map, and subsequent experiences are assigned a position based on the last experience's position and the agent's movement that has occurred since.

Each experience node has connections to a single local view cell, representing in our case the distinctive neural data observed when the rat is located somewhere within the maze. The local view cells acts as a controller for the experience map, activating and deactivating their corresponding experience nodes according to their own activation:
\begin{align}
    E_i = \left\{
  \begin{array}{@{}ll@{}}
    0, & \text{if}\ V_{\rm curr} \neq V_i, \\
    1, & \text{if}\ V_{\rm curr} = V_i,
  \end{array}\right.
\end{align}
where $E_i$ is the activation of the $i$th experience node, $V_{\rm curr}$ is the currently activated local view cell, and $V_i$ is the view cell associated with experience $i$. 

New experience nodes are created during the creation of their corresponding local view cells, 
when the input data is deemed dissimilar enough to data associated with other local view cells. Upon instantiation, experience nodes are assigned location according to the location represented by the current center of activation of the pose cell network.

\section{Results}
\label{sec:reults}

\subsection{Neural Decoding Results}
\label{ssec:decode-results}

\begin{figure}[ht]
   \centering
\begin{tabular}{cc}
\includegraphics[width=0.5\linewidth]{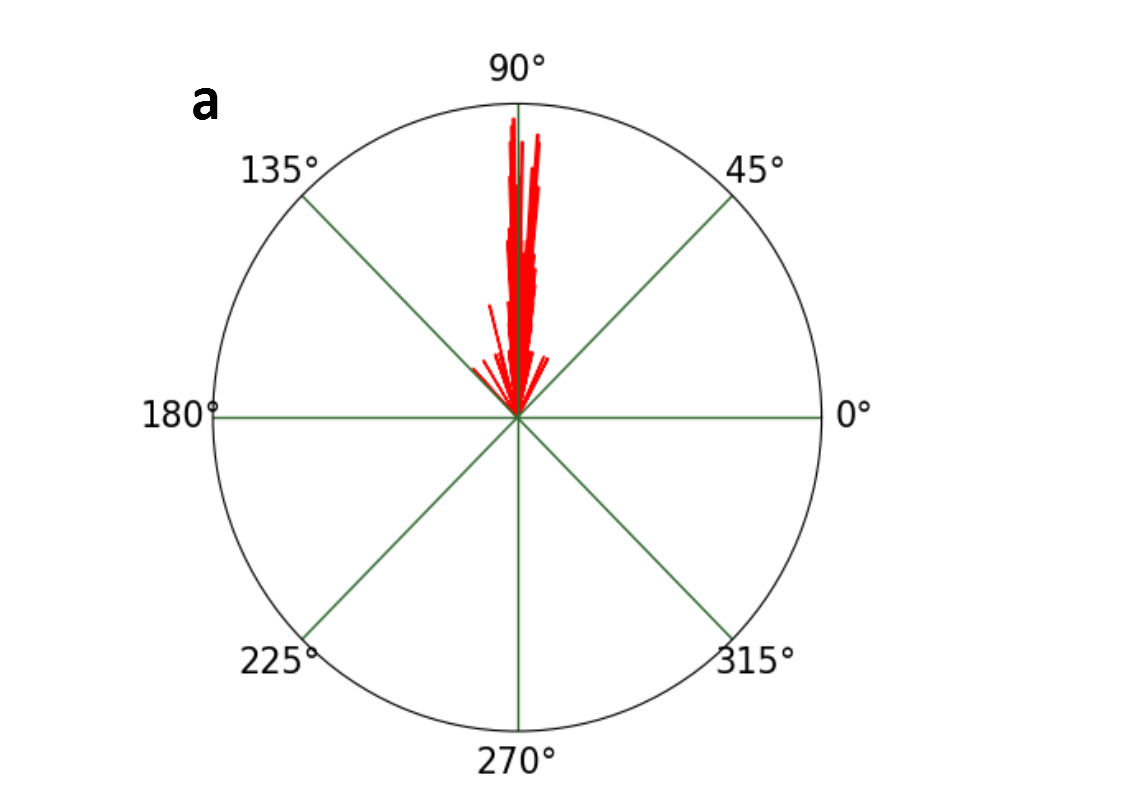} &
\includegraphics[width=0.5\linewidth]{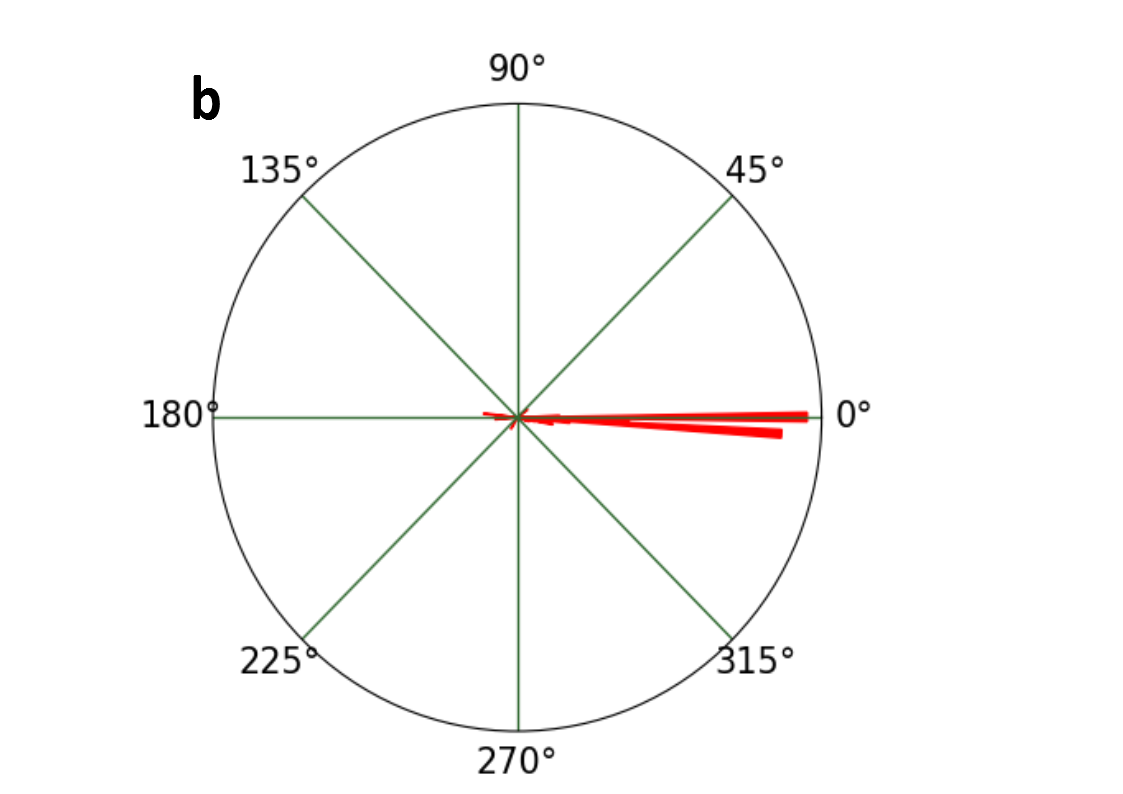} \\
\includegraphics[width=0.5\linewidth]{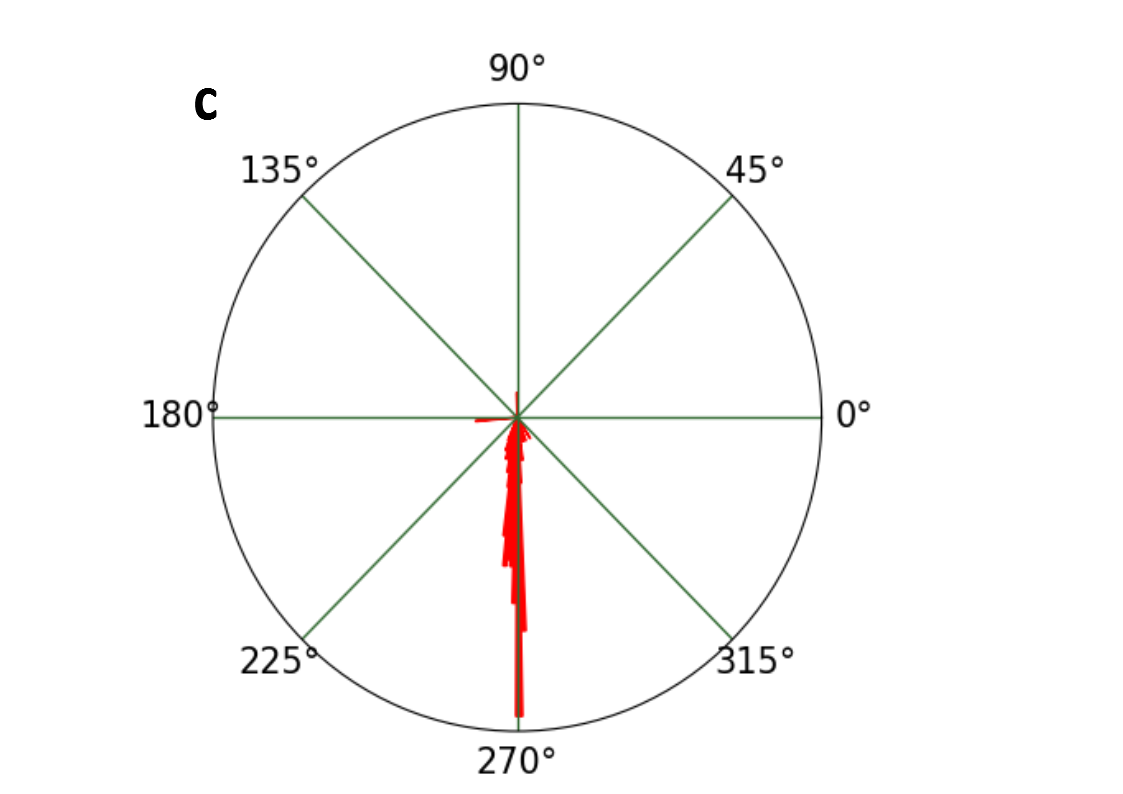} &
\includegraphics[width=0.5\linewidth]{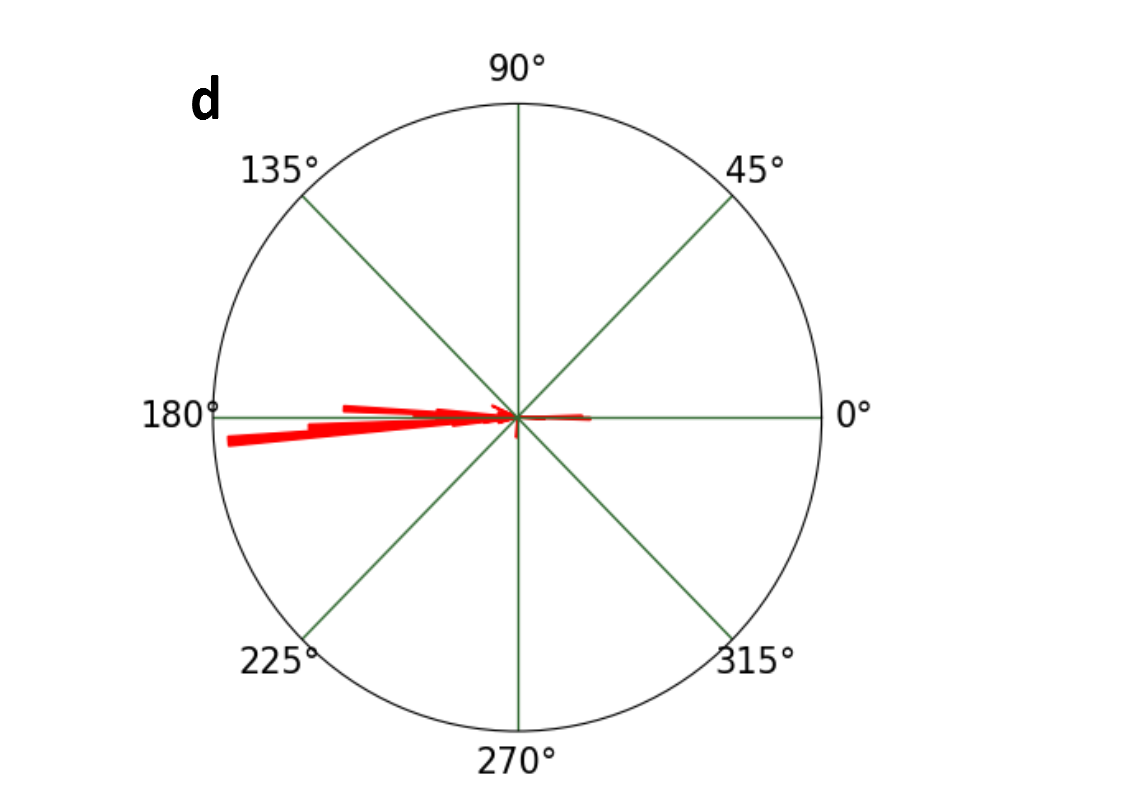}
\end{tabular}
    \caption{Angular histograms showing the distributions of estimated directions for each true direction value for one rat; (a) North, (b) East, (c) South, and (d) West. We see that the system is able to decode all directions with good accuracy. We note that perfect alignment with cardinal angles is not expected due to imperfections in the maze shape.}
    \label{fig:dirhists} 
\end{figure}

Experiments were conducted for three rats using data collected from one maze task session per rat, with data ranging from 28 to 34 minutes in duration. The initial 80\% of data was used to train the deep neural decoding system, while the final 20\% was used for testing purposes and to generate the decoded maps shown in Figure \ref{fig:decodedmaps}. Training models for each rat took an average of 13.5 hours using an NVIDIA 1070 GPU.

\begin{table}[ht]
\caption{Mean absolute location, direction, and speed error for all rats considered.}
\begin{tabular}{rlll}
  & \textit{Location (cm)} & \textit{Direction (deg)} & \textit{Speed (cm/s)} \\
  \hline 
Rat 1 & 2.188             & 7.816               & 0.486           \\
Rat 2 & 1.641             & 6.997               & 0.316           \\ 
Rat 3 & 1.849             & 12.354             & 1.487        
\end{tabular}
\label{tab:decodingresults}
\end{table}

High decoding accuracy was achieved for each rat, with mean absolute error (MAE) for location decoding at 2.188cm, 1.641cm, and 1.849cm, direction MAE at 7.816$^\circ$,  6.997$^\circ$, and 12.354$^\circ$, and speed MAE at 0.486cm/s, 0.316cm/s, and 1.487cm/s, respectively (shown in Table \ref{tab:decodingresults}). Notably, this is the first example of a deep convolutional neural network decoding direction of travel directly (as opposed to head direction, which is typical). As only four directions are possible within the considered maze, we show histograms in Figure \ref{fig:dirhists} visualising the distribution of estimated directions for each true direction value for one rat. We see that the system is able to decode all directions highly accurately, this is despite large class imbalances, with the `east' direction being over-represented in the training data for this rat.

\subsection{SLAM Results}
\label{ssec:slamres}

\begin{figure}[ht]
   \centering
\includegraphics[width=\linewidth]{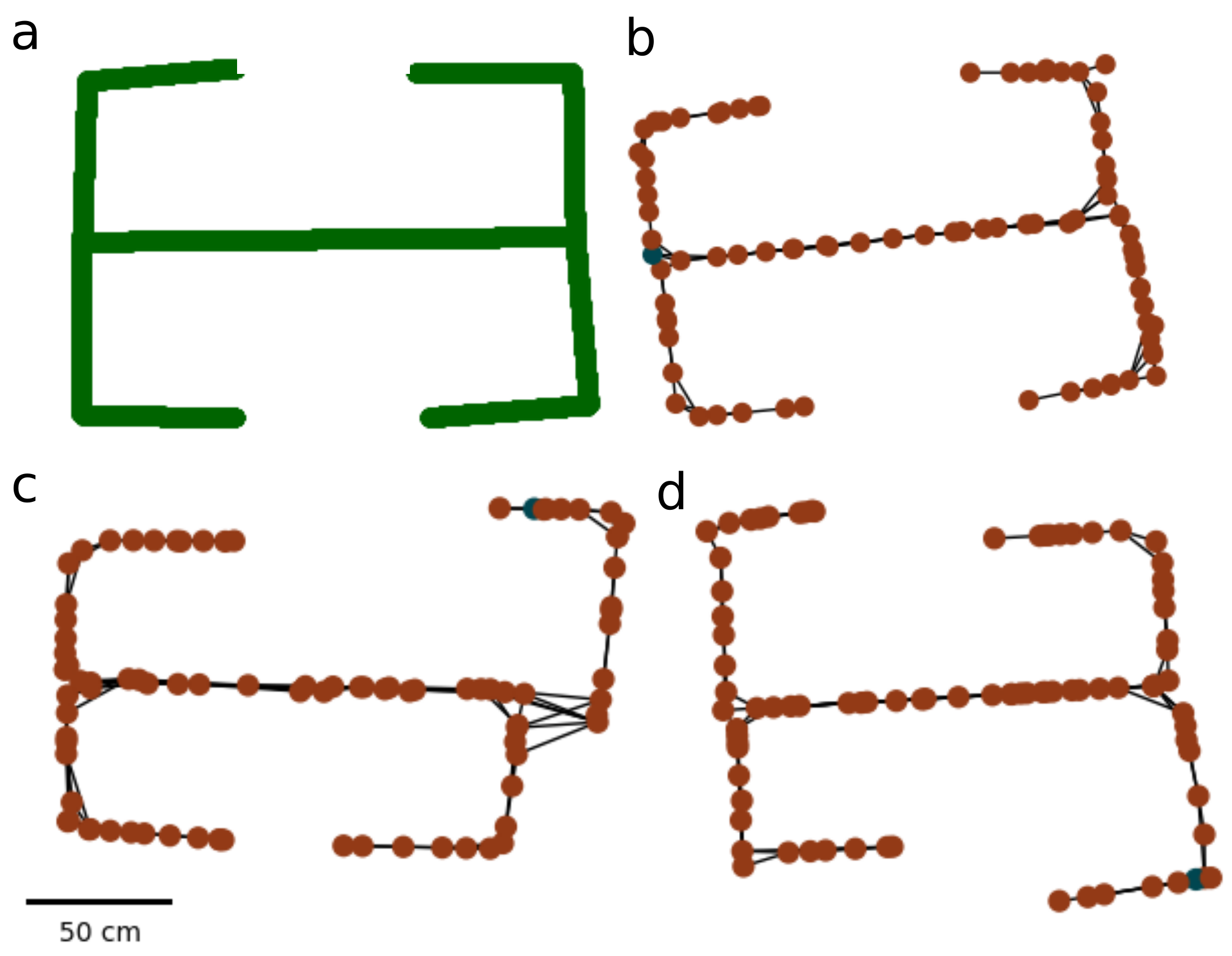}\\
    \caption{The true shape of the environment being explored by the rats (a), and inferred cognitive maps from each of the three rats as generated by the system presented here from $\sim 6$ minutes of test data (b,c,d). The shape and scale of the inferred maps was accurate in all cases. A photograph of the true maze is shown in Figure \ref{fig:bigfig} for comparison.}
    \label{fig:decodedmaps} 
\end{figure}

Using each rat's corresponding deep neural decoder, we performed the  RatSLAM algorithm on testing data to produce the decoded cognitive maps presented in Figure \ref{fig:decodedmaps}. We see that for all rats, the shape and scale of the maze was correctly inferred. We also note that there is very little aliasing in the system, indicating a functional and effective loop closure system; for comparison, maps generated with no functional loop closure module are presented in the supplemental material. Maps generated using inaccurate loop closure modules may also create erroneous links between distant and unrelated nodes, so it is encouraging to note that this has not happened for the maps shown in Figure \ref{fig:decodedmaps}.

We note that for one rat, a significant decoding error has occurred on the right decision point of the maze (Figure~3c). This error is due to the deep neural decoder producing an incorrect speed prediction during the first navigation through this point present in the testing data. As the algorithm presented here has no method for iteratively improving location estimates of experience nodes upon re-exploration, an incorrect prediction generated during initial exploration will remain in the generated maps. 

During creation of of these maps, an estimate of the rat's position within the environment is also maintained. The MAE of this location estimate for each rat was 18.353cm, 13.808cm, and 28.118cm, respectively. We note that though this error is higher than achieved by predicting location directly using the deep neural decoder (Table 1), our SLAM method may be extendable to predicting location even in novel environments, which would be previously unseen by both the deep neural decoder and the rodent. In contrast, predicting location directly with a deep network requires the network to first be trained using location training data from that same environment.

\section{Discussion \& Further work}
\label{sec:discussion}

This paper has provided an approach for decoding and visualising cognitive maps using intracranial neural activity data. All maps generated were high fidelity graphical representations of the environments being explored, capturing the correct shape and scale of the environment being explored.

It has been shown that head direction and speed decoding is partially invariant to environment \cite{taube1990head}. Thus, a natural avenue for further research would be to extend BrainSLAM to datasets with multiple environments; where training data is generated in one environment, and testing data in another. No such data was available for use in this project.

Other work should aim to incorporate an experience map correction process for minimizing the discrepancies between relative spatial information encoded in the edges between experience nodes, and the locations of those experience nodes in experience map space. For example, two experience nodes may be far away in the experience map but linked during a movement which our odometry module calculates to be a much smaller distance. In this case, by minimizing the discrepancies between the relative locations of experiences in the experience map and the inter-experience spatial transition information, the experience map could converge over time to a higher fidelity representation of the map being explored \cite{milford2005experience, milford2007learning}

Further work should also explore to what extent the maps generated by the system presented here can be used to predict a rats behaviours as it moves through the maze. If it can be shown that errors in the generated map are associated with related sub-optimal navigation decisions in the maze, this is evidence that this system is generating maps which are in some way congruent to the true internal cognitive map of the animal. As an example, consider an experiment showing that a rat with will take a route which is perceived as shorter in the generated cognitive map of the maze, despite it being the same length as alternative routes in reality. 

More recent research into hippocampal function \cite{solomon2019hippocampal}, suggests that hippocampal activity tracks distance in semantic space as well as physical. This supports the growing consensus that the hippocampus is partially responsible for the maintenance of a domain-general cognitive map \cite{spiers2020hippocampal}. Thus the research presented here could not only facilitate the use of biological agents to map physical spaces, but also to map more abstract semantic spaces.

\section{Conclusion}

Using a novel combination of RatSLAM and a deep learning approach to decoding behavioural variables from wide-band neural activity, this paper has presented the first approach to decoding and visualising cognitive maps using only intracranial local field potential data. Maps have been generated using data from the brains of three rats, with data being gathered from the dorsal CA1 of hippocampus, prefrontal cortex, and the parietal cortex. All maps generated were high fidelity graphical representations of the environments being explored, capturing the correct shape and scale of the environment being explored. While maps were being inferred, the system was simultaneously able to maintain reasonable location estimates for all rats. 

The ability to visualise and quantify the cognitive map of an animal opens up new avenues of research into the role of these these maps in navigation and decision making. We believe this work has value in the field of bio-inspired control; shedding light on the ways that animals form and utilize maps of environments, and perform navigation in a complex world. 
\balance
Further, the research presented here extends SLAM algorithms to utilize a novel modality; neural local field potentials. This could facilitate a wide variety of further applications related to brain computer interfaces, automated mapping using biological agents, and environment exploration. 

\begin{acks}
We thank the EPSRC, BBSRC (BB/G006687/1), Wellcome Trust (202810/Z/16/Z), MRC and Leverhulme Trust for support. For the purpose of open access, the author has applied a CC BY public copyright licence to any Author Accepted Manuscript version arising from this submission.
\end{acks}

\bibliographystyle{ACM-Reference-Format} 
\bibliography{refs}


\end{document}